\documentclass[10pt,twocolumn,letterpaper]{article}

\usepackage{cvpr}
\usepackage{times}
\usepackage{epsfig}
\usepackage{graphicx}
\usepackage{amsmath}
\usepackage{amssymb}

\usepackage[ruled]{algorithm2e}
\usepackage[caption=false]{subfig}
\usepackage{booktabs}
\usepackage[figuresright]{rotating}

\usepackage[pagebackref=true,breaklinks=true,letterpaper=true,colorlinks,bookmarks=false]{hyperref}

\cvprfinalcopy 

\def\cvprPaperID{6} 

\ifcvprfinal\fi
\begin{document}

\title{Stingray Detection of Aerial Images Using Augmented Training Images Generated by A Conditional Generative Model}

\author{Yi-Min Chou$^{1,2}$~~~~~Chien-Hung Chen$^{1,2}$~~~~~Keng-Hao Liu$^3$~~~~~Chu-Song Chen$^{1,2}$\\
$^1$ Institute of Information Science, Academia Sinica, Taipei, Taiwan\\$^2$ MOST Joint Research Center for AI Technology and All Vista Healthcare\\$^3$ Department of Mechanical Electro-mechanical Engineering, National Sun Yat-sen University,\\
Kaohsiung, Taiwan
\\
\tt\small \{chou, redsword26, song\}@iis.sinica.edu.tw,
\tt\small keng3@mail.nsysu.edu.tw,
}
\pagenumbering{gobble}
\maketitle

\begin{abstract}
In this paper, we present an object detection method that tackles the stingray detection problem based on aerial images.
In this problem, the images are aerially captured on a sea-surface area by using an Unmanned Aerial Vehicle (UAV), and the stingrays swimming under (but close to) the sea surface are the target we want to detect and locate.
To this end, we use a deep object detection method, faster RCNN, to train a stingray detector based on a limited training set of images.
To boost the performance, we develop a new generative approach, conditional GLO, to increase the training samples of stingray, which is an extension of the Generative Latent Optimization (GLO) approach.
Unlike traditional data augmentation methods that generate new data only for image classification, our proposed method that
mixes foreground and background together can generate new data for an object detection task, and thus improve the training efficacy of a CNN detector.
Experimental results show that satisfiable performance can be obtained by using our approach on stingray detection in aerial images.
\end{abstract}

\section{Introduction}
\label{sec:intro}

Detecting specific animals in aerial images captured by an UAV is a crucial research topic.
In this research direction, computer vision techniques are beneficial to the development of popular tools for biological researches.
In this paper, a stingray detection approach is introduced.
Stingrays are common in coastal tropical and subtropical marine waters.
They usually appear in surface water so that a common UAV can capture them.
In this work, the scenario we focus on is the automatic detection of stingray from the aerial images recorded on a sea-surface
area.

To monitor the behaviors and understand the distribution of a certain animal, biologists collect aerial photos or videos by an
UAV.
After obtaining the materials, they have to manually annotate the position, number, and size of the target animal from the image
scene.
This step is extremely tedious and time-consuming.
Besides, the collected photos could be partially useless because the target animal may be missing in the scene.
Therefore, using an automatic, computer-based method to recognize the target animal is necessary for the kind of research.

\begin{figure*}
	\begin{center}
	\includegraphics[width=0.85\textwidth]{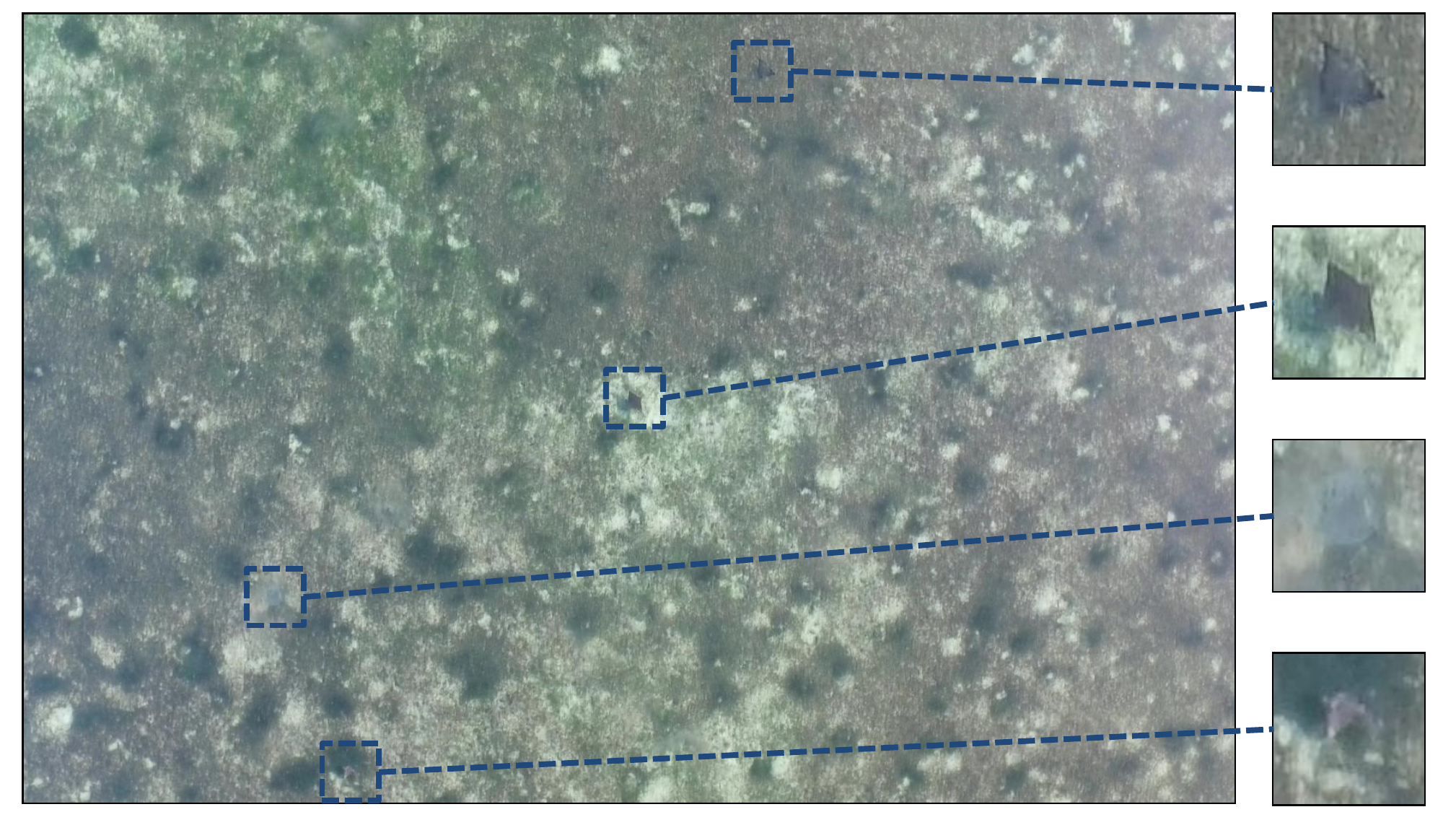}
	\end{center}
	\caption{A typical aerial sea-surface image, which contains four stingrays in the scene. The stingray detection problem is demanding because of the similar rocks/reefs under the water and the aerial images are filled with light reflection of water ripples.}
	\label{fig:example}
\end{figure*}

However, automaticly recognizing the stingray is demanding due to the following issues.
First, the color of stingrays is similar to that of the rocks/reefs under the water, and the stingrays could be occluded by the dust when swimming.
Second, the aerial images are usually filled with light reflection of water ripples;
Third, the shape of stingray is not always consistent, and is hard to define.
Under such circumstances, traditional machine learning methods accompanied with hand-craft features often fail for the detection
task on the sea-surface images aerially taken.
Figure~\ref{fig:example} demonstrates the difficulty of this problem.

As the rapid progress of deep learning (DL), it has been a popular approach to many image classification and object detection
tasks.
In recent years, a breakthrough of image recognition has been made via deep convolution neural networks
(CNN)~\cite{krizhevsky2012imagenet}.
Deep CNN enforces end-to-end training, so that feature extraction and classification are integrated in a single framework.
Besides handing the case where only one concept is contained in an image~\cite{krizhevsky2012imagenet, simonyan2014very, he2016deep}, deep
CNN has been extended for object detection~\cite{ren2017faster, liu2016ssd, redmon2017yolo9000}, where not only the objects
contained in an image are recognized but their sites are marked by tight bounding boxes.
In our case, the problem to be tackled belongs to 2-class object detection, where the foreground (or positive) class consists of
stingrays and the background (or negative) class consists of sea-surface patches.
We employ deep CNN detectors to fulfill our goal, where faster RCNN~\cite{ren2017faster} is used in our work.
The performance obtained is far more satisfied than that of using hand-craft features in our experience.
 



Nevertheless, DL usually requires a large set of training samples to learn the network weights, while the biological image materials
are sometimes insufficient to fulfill the demand.
There are two main difficulties encountered when using deep-learning object detector in our work.

\begin{itemize}
\item{Insufficient training data}:
The amount of training data is limited by the few number of UAV flights, and the image quality is inconsistent by weather condition,
environmental change, capturing location and latitude.
It results in the lack of effective training images and data diversity.

\item{Background transparency}:
Because the sea surface is translucent, the stingray image is actually embedded into water but not explanted on the water.
Thus, the color of stingray is blended that of water.
The conventional data augmentation approaches could not generate this type of images.
\end{itemize}

Our stingray detection problem has 2 classes (foreground and background).
To tackle this problem, we introduce a mixed background and foreground (bg-fg) data augmentation approach to handle the
problem.
Our approach, namely, conditional GLO (C-GLO), can learn a generator network that produces a foreground object given a
specified background patch.
C-GLO can learn the distribution of foreground (w/ stingray) and background (w/o stingray) images
simultaneously in the latent space with a single network.

Once the generator is learned, we can freely generate the synthetic stingray images respect to any sea-surface background to enrich
the amount and the diversity of the training dataset.
For the detection, we used Faster R-CNN as the CNN-detector for the evaluation. 
The experimental results show that using the C-GLO augmented samples for training can satisfiedly improve the detection performance.
Such an augmentation approach could be potentially applied to other analogous applications.

\begin{figure*}
	\begin{center}
	\includegraphics[width=0.6\textwidth]{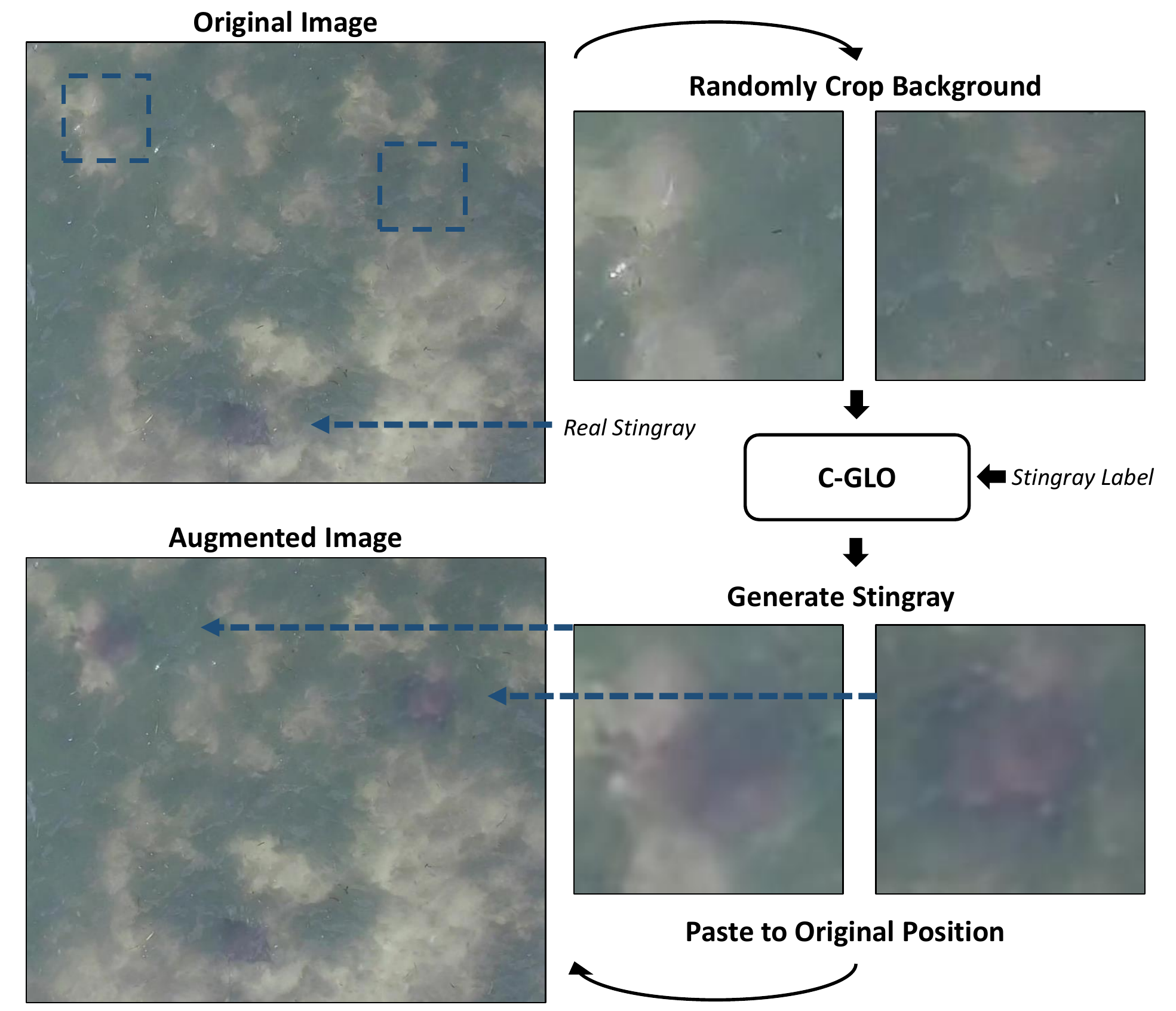}
	\end{center}
	\caption{Mixed Bg-Fg Syntheses. Given sea-surface patches cropped in the original training images, we generate a stingray inside each patch and put the patch back to the original image. The augmented images obtained therefore contain more stingrays. In this way, the training set of images is re-generated such that each image has sufficient many stingrays and the number of stingrays per image is approximately the same.}
	\label{fig:mixfgbg}
\end{figure*}

\section{Related Work}
In this section, we briefly review works related to our study on two folds: deep-CNN object detection and generative networks.

\subsection{Deep CNNs for Object Detection}\label{sec:reviewobjdetc}
Object detection methods have been made great progress recently with the resurgence of CNNs.
In the past, researches focus on the design of useful hand-crafted features, such as HOG and DPM.
Currently, it shifts to the design of a good CNN architecture that can automatically capture high-level features for detection.

DL-based detection approaches started with R-CNN~\cite{girshick2016region} that adopts an additional selective search procedure. Later, this kind of method evolved to an approximate end-to-end model with using reginal proposal network (RPN) in Faster R-CNN~\cite{ren2017faster}.
Many follow-up studies successively improve the performance such as R-FCN\cite{dai2016r} and Mask R-CNN~\cite{he2017mask}, or accelerate the computation such as SSD~\cite{liu2016ssd} and YOLO2~\cite{redmon2017yolo9000}.


\subsection{Generative Models}\label{sec:reviewgennet}
Nature images generation has been investigated by the work of Variational Autoencoders (VAE) \cite{kingma2013auto}.
Later, Goodfellow et al.~proposed Generative Adversarial Networks (GAN) \cite{goodfellow2014generative} that trains a generator and a discriminator simultaneously via an iteratively adversarial process.
GAN has demonstrated the capability of generating more convincing images than VAE. 

Although GANs provide sharper images, a main drawback lies in the difficulty of converging to an equilibrium state during training.
Recently, numerous GAN-related studies have been proposed
\cite{radford2015unsupervised,arjovsky2017wasserstein,berthelot2017began,salimans2017improved}, and most
of them focus on resolving the problems of model instability and mode collapse
\cite{dumoulin2016adversarially,metz2016unrolled,arjovsky2017towards}.
Nevertheless, training of GAN is still more demanding and relatively unstable compared to pure supervised training.

To avoid challenging adversarial training protocol in GAN, Bojanowski et al.~proposed Generative Latent Optimization (GLO)~\cite{GLO}.
GLO removes the discriminator in GAN and learns the mapping from images to noise vectors by minimizing the reconstruction loss. 
It provides a stable training process while enjoys many of the desirable properties of GAN, such as synthesizing appealing images and interpolating meaningfully between samples.

\begin{figure*}
	\begin{center}
	\includegraphics[width=0.8\textwidth]{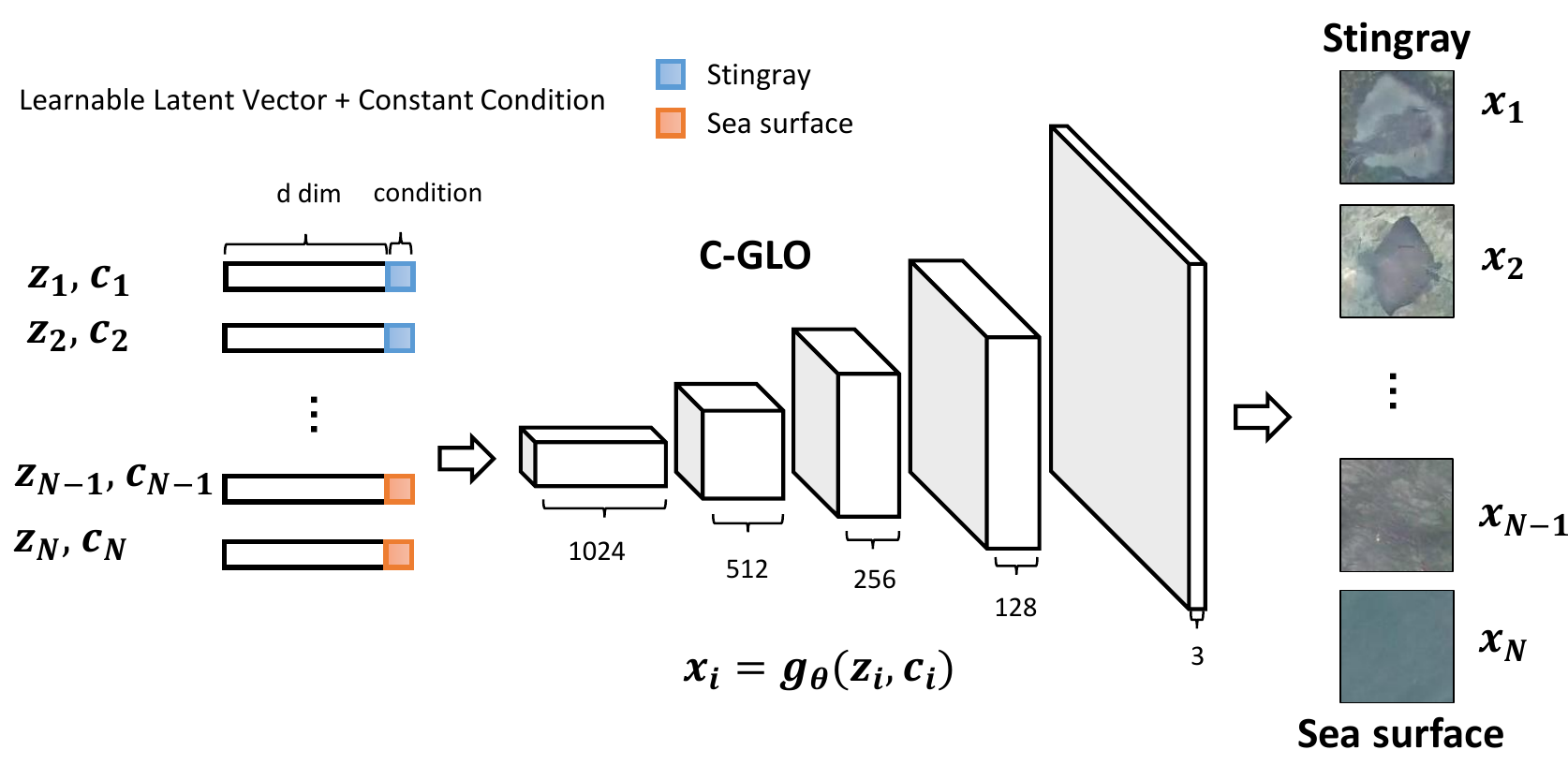}
	\end{center}
	\caption{Conditional GLO (C-GLO) introduced in this work.}
	\label{fig:CGLO}
\end{figure*}

\section{Our Method}
Data augmentation (such as cropping and flipping the images) has been widely used for the training of image classifiers, where the labels are provided for the entire image.
However, the task of object detection requires bounding-box outputs, while augmenting the training images with
bounding-box samples of the objects is more difficult.

In object detection, the positive patches are often far fewer than the negative ones.
For example, in our data, sometimes only one stingray is contained in a training image, which makes a CNN detector demanding to train.
We introduce a method that performs data augmentation in the learning phase for object detection.
Considering that the sea surface is translucent, we propose to use a generator that produces foreground objects mixed with the background patches selected from the image.
Given some background (i.e., sea-surface) patches randomly cropped from the original image (as shown in the upper half of
Figure~\ref{fig:mixfgbg}), we use the C-GLO approach to synthesize a foreground object (i.e., stingray) per each background patch,
and put them back to the original sites in the image (as shown in the lower half of Figure~\ref{fig:mixfgbg}).


In the following, we introduce C-GLO at first in Section \ref{sec:cglo}, and then the mixed bg-fg synthesis and the CNN detector
in Section \ref{sec:objdet}.

\subsection{Conditional GLO and Architecture Adopted}
\label{sec:cglo}

GLO~\cite{GLO} is a generative method introduced by Bojanowski et al.
Given unsupervised training images $\mathbf{I}=\{I_1, \cdots, I_N\}$, GLO trains a generator $\mathbf{\Phi}$ (with the input $z$ and network weights $W$), such that the following objective is minimized:
\begin{equation}
e(W,\mathbf{z})=\sum_{i=1}^{N} loss(\mathbf{\Phi}(W;z_i) - I_i),
\end{equation}
where $\mathbf{z}=\{z_1, z_2, \cdots, z_N\}$.
A two-stage iterative method is introduced for the minimization:

\begin{enumerate}
\item Fixing $\mathbf{z}$, find $W$ to reduce $e(W,\mathbf{z})$ via back-propagation;

\item Fixing $W$, find $z_i$ to reduce $loss(\mathbf{\Phi}(W;z_i) - I_i)$ via back-propagation, $\forall i$, with an uni-model normalization to $\mathbf{z}$.
\end{enumerate}

\noindent The above two steps are iterated to refine $W$ and $\mathbf{z}$ alternatively. GLO holds the following advantages.

\begin{itemize}

\item{Direct training}:
First, GLO learns a generative network directly with no needs of other complemented networks.
In GAN, a discriminant networks is further used to form a two-player game for the generator leraning.
However, GANs easily suffer from the problem of instable training.
Though many modification of GANs~\cite{arjovsky2017wasserstein,berthelot2017began,salimans2017improved} have been proposed to address this issue, the training process of GANs is still relatively unstable compared to supervised training.
On the contrary, GLO's training process is more alike supervised training and thus it is easier to get stable results in our
experience.
Besides GAN, VAE also requires an additional encoder for the generator training.
GLO can train the generator directly and thus consumes fewer training resources.

\item{Inverse mapping}:
A second advantage of GLO is its reconstruction capability.
Assume that the generator has been trained, and thus $W$ is known.
Given an image $I_i$, the latent codes $z_i$ that exactly generates $I_i$ can be found via iterating step 2 of the above training process (with $W$ fixed).
Hence, the inverse mapping of the image $I_i$ is available, which is unlike GAN that can generate novel images but do not provide the codes that recover the original images.
Although some approaches combining GAN and autoencoder (such as~\cite{berthelot2017began}) can find the latent codes via the encoder subnetwork of the autoencoder, the codes are obtained via a forward mapping indirectly and thus the recovery performance is not guaranteed.
With the reconstruction capability, given an image patch cropped from sea surface, GLO can thus find the latent code $z$ that produces the same patch that can be seamlessly put back to the sea surface, which suits our data-augmentation approach introduced later.
\end{itemize}

\begin{figure*}
	\begin{center}
	\includegraphics[width=0.7\textwidth]{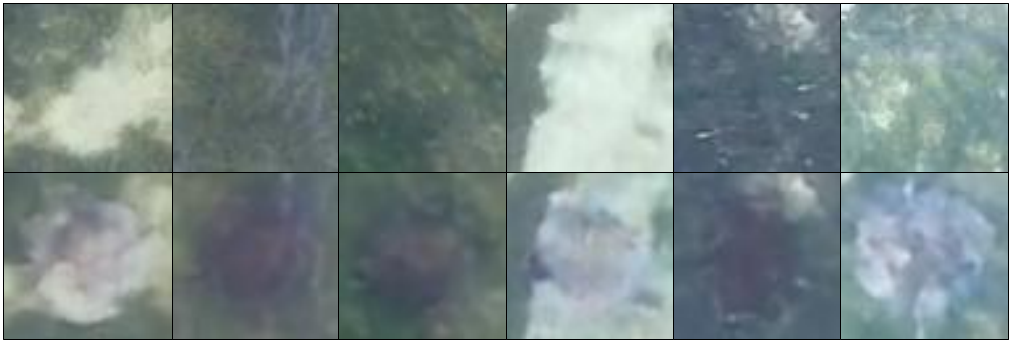}
	\end{center}
	\caption{Switch the condition in the latent space to convert a background patch to a mixed bg-fg patch via C-GLO with the dimension of latent code $d$ = 256; upper: the original background patch; bottom: the synthetic patch.}
	\label{fig:switch}
\end{figure*}

We extend GLO to C-GLO as follows.
Unlike GLO, the latent space input is generalized to $(z,c)$ in C-GLO, where $z\in R^d$ is the latent code and $c\in R^m$ is a set of ``on-off" labels.
In this study, $m=1$ since only a single condition (Fg or Bg) is required.
The training images thus become $\mathbf{c}=\{(I_1,c_1), \cdots, (I_N,c_N)\}$, where $c_i\in \{0,1\}$ represents background and foreground, respectively.

The training process of C-GLO is similar to that of GLO as follows:
\begin{enumerate}
\item Given $\mathbf{z},\mathbf{c}$, find $W$ to reduce the total reconstruction loss of $\mathbf{I}$. 
\item Given $W,c_i$, find $z_i$ to reduce the reconstruction loss of $I_i, \forall i$. 
\end{enumerate}
\noindent The above two steps are executed iteratively. 

Figure~\ref{fig:CGLO} shows the architecture of the C-GLO adopted.
Without loss of generality, we use the same de-convolution network in DCGAN~\cite{radford2015unsupervised} as the architecture of our C-GLO in this work.
C-GLO inherits the characteristics of GLO: easy to train and provides explicit latent codes for image reconstruction.
The learned C-GLO can then be used to generate novel images of stingray (or sea-surface) via the condition $c=1$ (or $c=0$) and the respective codes of $z$.

\subsection{Mixed Bg-Fg Syntheses and object detector}
\label{sec:objdet}

To convert a given background patch to a mixed bg-fg one, we disentangle the condition label of the latent representation.
Let $\mathbf{\Phi}$ be a trained generator (with the weights $W$).
Consider a background ($c=0$) image patch, say $I_{i_b}$; let $z_{i_b}$ be its inverse mapping (i.e., $\mathbf{\Phi}(W;z_{i_b};c) = I_{i_b}$).
We then switch the condition label from $c=0$ to $c=1$ and keep the other parameters $W, z_{i_b}$ unchanged.
By doing so, the sea surface patch specified by the latent code $z_{i_b}$ is provided with a positive condition $c=1$.
It results in the effect that the sea surface patch $I_{i_b}$ contains a stingray image inside it.
The disentangled patch (with a synthesized stingray in it) can thus be put back to the entire sea-surface scene without noticeable artifact.
The sea surface image are then augmented with more stingrays for training.
Figure~\ref{fig:mixfgbg} gives some examples of the augmented samples.
More examples can be found in Figure~\ref{fig:switch}.

\begin{figure*}
	\begin{center}
	\includegraphics[width=0.75\textwidth]{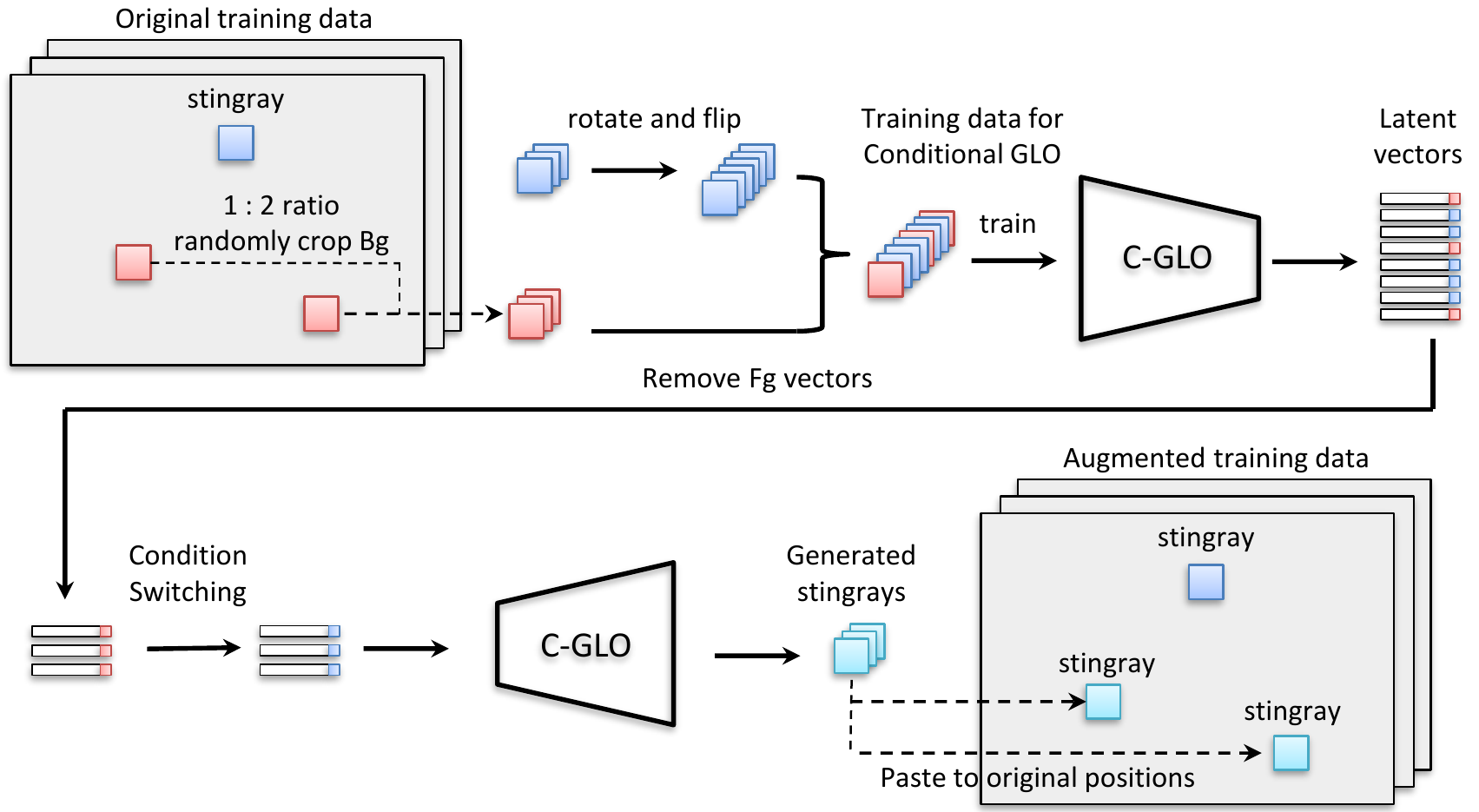}
	\end{center}
	\caption{An overview of our approach.}
	\label{fig:overview}
\end{figure*}

We apply the augmented data to train an existing CNN detector, Faster R-CNN.
Faster R-CNN contains three parts of networks, the feature-extration network, region proposal network, and classification network.
The architecture of the feature-extraction network can be flexibly chosen.
In this work, we use two network models, ZF model~\cite{ZFNet} and VGG-16 model~\cite{simonyan2014very}, as the architecture of the feature-extraction network.
The two Faster R-CNNs are evaluated on our dataset with or without our data augmentation method for comparison.
An overview of our approach is given in Figure~\ref{fig:overview}.






\section{Experiments}
In this section, we apply our mixed bg-fg synthesis approach to stingray detection and present the results.

\subsection{Dataset and Experimental Settings}
We have gotten a total of 36 labeled videos taken in the day time, recorded at 4k (3840$\times$2160) resolution.
The stingray images are sampled from the videos at 1fps or 4fps.
Those images are composed of various components such as rocks, ripples, dust, and light reflections, and thus the stingrays are difficult to be detected even by human.
We select 3245 images (from 16 videos) for training and 3147 images (from the rest 20 videos) for testing.
All the images are re-scaled to 
1920$\times$1080 for learning because of the limited GPU memory (a single Nvidia Titan-X GPU is used in our experiment).
In those images, there is only one object class (stingray) with the size within  
30 to 350 pixels. 
Hence, the anchor-box parameters in Faster R-CNN are set to reflect the scales accordingly, while
the other settings follow the default 
of Faster R-CNN.
To train the C-GLO model, $L_1$ loss is used and the output size is 64$\times$64 pixels.
For each training image, we crop the stingray patches as the positive samples and randomly crop sea-surface patches as the negative ones.
%
After further augmentation via rotation and flipping of the stingrays, we finally use 30496 stingray patches and 7664 sea-surface patches for training the C-GLO model.



\subsection{Data Augmentation Results}
We switch the condition of the trained GLO to generated the mixed bg-fg patches for data augmentation, as described in Section~\ref{sec:objdet}.
Figure~\ref{fig:switch} shows several of our mixed bg-fg synthetic patches.
It can be seen that our method can generate new stingray of various colors and shapes while keeping the same surroundings of the original patches.


\subsection{Detection Results}
We expect the detection capability of Faster R-CNN can be benefited from the data augmented by C-GLO.
The detection results are reported in Table 1.
It can be seen that the Average Precision (AP) can be improved by 4.15 and 2.02 percents when using ZF and VGG-16 as the base models for feature extraction in Faster R-CNN, respectively.
In addition, there is only a slight difference on the performance by changing the dimension of $z$.
It reveals that the detection capability of our approach is insensitive to the size of the latent space.
Also, our approach is capable of generating diverse patches to augment the training dataset, which enforces a more effective training of object detectors and improves the performance.

\begin{table}[pt]
    \centering
	\caption{Average precision (AP) obtained via our augmentation method ($d$ = 128, 256, 512) compared to the  Faster R-CNN without augmentation}
	\label{LeNetMergeOrSingle}
	\begin{tabular}{lcccc}
	    \\\hline
		Network       &Baseline         & Ours-128  & Ours-256            & Ours-512 \\ \hline
		ZF            & 78.89     & 82.75     & 82.42            & \textbf{83.04}  \\ \hline
		VGG-16           & 84.59      & 86.14    & \textbf{86.61}   & 86.43           \\ \hline	
	\end{tabular}
\end{table}

\section{Conclusions}
In this paper, we present a method to detect stingrays in aerial images. 
We introduce a data augmentation method called \emph{mixed bg-fg synthesis} to fuse background patches and foreground objects without apparent artifacts, which is achieved by a new generative network C-GLO. 
The experimental results 
reveal that the object detection performance can be improved via our data augmentation method. 
The system developed in this work can help biologists to track and annotate stingrays automatically.

Currently, our approach is based on images.
In the future, we plan to extend our approach to video-based data augmentation and objection detection.

\section*{Acknowledgements}
This work is supported in part by the projects MOST 107-2634-F-001-004 and MOST 106-2221-E-110-074.

{\small
\bibliographystyle{ieee}
\bibliography{deepobj}

\begin{thebibliography}{10}\itemsep=-1pt

\bibitem{arjovsky2017towards}
M.~Arjovsky and L.~Bottou.
\newblock Towards principled methods for training generative adversarial
  networks.
\newblock In {\em ICLR}, 2017.

\bibitem{arjovsky2017wasserstein}
M.~Arjovsky, S.~Chintala, and L.~Bottou.
\newblock Wasserstein gan.
\newblock {\em arXiv preprint arXiv:1701.07875}, 2017.


\bibitem{berthelot2017began}
D.~Berthelot, T.~Schumm, and L.~Metz.
\newblock Began: Boundary equilibrium generative adversarial networks.
\newblock {\em arXiv preprint arXiv:1703.10717}, 2017.

\bibitem{GLO}
P.~Bojanowski, A.~Joulin, D.~Lopez-Paz, and A.~Szlam.
\newblock Optimizing the latent space of generative networks.
\newblock {\em arXiv preprint arXiv:1707.05776}, 2017.

\bibitem{dai2016r}
J.~Dai, Y.~Li, K.~He, and J.~Sun.
\newblock R-fcn: Object detection via region-based fully convolutional
  networks.
\newblock In {\em Advances in neural information processing systems}, pages
  379--387, 2016.

\bibitem{dumoulin2016adversarially}
V.~Dumoulin, I.~Belghazi, B.~Poole, A.~Lamb, M.~Arjovsky, O.~Mastropietro, and
  A.~Courville.
\newblock Adversarially learned inference.
\newblock {\em arXiv preprint arXiv:1606.00704}, 2016.

\bibitem{girshick2016region}
R.~Girshick, J.~Donahue, T.~Darrell, and J.~Malik.
\newblock Region-based convolutional networks for accurate object detection and
  segmentation.
\newblock {\em IEEE transactions on pattern analysis and machine intelligence},
  38(1):142--158, 2016.

\bibitem{goodfellow2014generative}
I.~Goodfellow, J.~Pouget-Abadie, M.~Mirza, B.~Xu, D.~Warde-Farley, S.~Ozair,
  A.~Courville, and Y.~Bengio.
\newblock Generative adversarial nets.
\newblock In {\em NIPS}, 2014.

\bibitem{he2017mask}
K.~He, G.~Gkioxari, P.~Doll{\'a}r, and R.~Girshick.
\newblock Mask r-cnn.
\newblock In {\em Computer Vision (ICCV), 2017 IEEE International Conference
  on}, pages 2980--2988. IEEE, 2017.

\bibitem{he2016deep}
K.~He, X.~Zhang, S.~Ren, and J.~Sun.
\newblock Deep residual learning for image recognition.
\newblock In {\em Proceedings of the IEEE conference on computer vision and
  pattern recognition}, pages 770--778, 2016.

\bibitem{kingma2013auto}
D.~P. Kingma and M.~Welling.
\newblock Auto-encoding variational bayes.
\newblock In {\em ICLR}, 2014.

\bibitem{krizhevsky2012imagenet}
A.~Krizhevsky, I.~Sutskever, and G.~E. Hinton.
\newblock Imagenet classification with deep convolutional neural networks.
\newblock In {\em NIPS}, 2012.

\bibitem{liu2016ssd}
W.~Liu, D.~Anguelov, D.~Erhan, C.~Szegedy, S.~Reed, C.-Y. Fu, and A.~C. Berg.
\newblock Ssd: Single shot multibox detector.
\newblock In {\em European conference on computer vision}, pages 21--37.
  Springer, 2016.

\bibitem{metz2016unrolled}
L.~Metz, B.~Poole, D.~Pfau, and J.~Sohl-Dickstein.
\newblock Unrolled generative adversarial networks.
\newblock In {\em ICLR}, 2017.

\bibitem{radford2015unsupervised}
A.~Radford, L.~Metz, and S.~Chintala.
\newblock Unsupervised representation learning with deep convolutional
  generative adversarial networks.
\newblock In {\em ICLR}, 2016.

\bibitem{redmon2017yolo9000}
J.~Redmon and A.~Farhadi.
\newblock Yolo9000: Better, faster, stronger.
\newblock In {\em Computer Vision and Pattern Recognition (CVPR), 2017 IEEE
  Conference on}, pages 6517--6525. IEEE, 2017.

\bibitem{ren2017faster}
S.~Ren, K.~He, R.~Girshick, and J.~Sun.
\newblock Faster r-cnn: towards real-time object detection with region proposal
  networks.
\newblock {\em IEEE transactions on pattern analysis and machine intelligence},
  39(6):1137--1149, 2017.

\bibitem{salimans2017improved}
T.~Salimans, I.~Goodfellow, W.~Zaremba, V.~Cheung, A.~Radford, and X.~Chen.
\newblock Improved techniques for training gans.
\newblock In {\em NIPS}, 2017.

\bibitem{simonyan2014very}
K.~Simonyan and A.~Zisserman.
\newblock Very deep convolutional networks for large-scale image recognition.
\newblock In {\em ICLR}, 2015.

\bibitem{ZFNet}
M.~D. Zeiler and R.~Fergus.
\newblock Visualizing and understanding convolutional networks.
\newblock In {\em Proceedings of the European Conference on Computer Vision},
  pages 818--833, 2014.

\end{thebibliography}
}

\end{document}